\documentclass[journal,12pt,onecolumn]{IEEEtran}
\usepackage{comment}
\usepackage{amssymb, amsmath,graphicx,charter, latexsym}
\usepackage{amsfonts}
\usepackage{bbm}
\usepackage{algorithm}

\newtheorem{theorem}{Theorem}
\newtheorem{remark}{Remark}
\newtheorem{assumption}{Assumption}
\usepackage{graphicx}
\usepackage{url}
\usepackage{epstopdf}
\usepackage{gensymb}
\IEEEoverridecommandlockouts                              

\newcommand{\ust}{^{\star}}

\newcommand{\cX}{\mathcal{X}}
\newcommand{\cF}{\mathcal{F}}

\newcommand{\bP}{\mathbb{P}}

\newcommand{\cV}{\mathcal{V}}

\newcommand{\cI}{\mathcal{I}}

\newcommand{\cE}{\mathcal{E}}
\newcommand{\bE}{\mathbb{E}}

\begin{document}
\title{A Partially Observable MDP Approach for Sequential Testing for Infectious Diseases such as COVID-19}

\author{Rahul Singh, Fang Liu, and Ness B. Shroff
\thanks{R. Singh is at the Department of Electrical Communication Engineering, Indian Institute of Science, Bengaluru; Fang Liu and Ness B. Shroff are at the Department of ECE and CSE, The Ohio State University. 

{\tt\small rahulsingh@iisc.ac.in, liu.3977@buckeyemail.osu.edu, shroff.11@osu.edu.}
}
}

\maketitle
\IEEEpeerreviewmaketitle

\begin{abstract}
The outbreak of the novel coronavirus (COVID-19) is unfolding as a major international crisis whose
influence extends to every aspect of our daily lives. Testing is critical in identifying patients and carriers.
Effective testing allows infected individuals to be quarantined, thus reducing the spread of COVID-19,
saving countless lives, and helping to restart the economy safely and securely. The testing capacity will
remain a constraint for the foreseeable future compared to the size of the US population. The focus of this paper is to develop highly efficient testing strategies that make optimal use of our testing resources in order to minimize the number of infected individuals.

Developing a good testing strategy can be greatly aided by contact tracing that provides health care
providers information about the whereabouts of infected patients in order to determine whom to test.
There have been significant efforts to improve contact tracing by developing apps that leverage the
ubiquity of smart phones to automatically detect contacts between individuals within a pre-determined
distance from each other (e.g., within 6 feet), the time duration of the contact, etc. Countries that have
been more successful in corralling the virus typically use a ``test, treat, trace, test'' strategy that begins
with testing individuals with symptoms, traces contacts of positively tested individuals via a combinations
of patient memory, apps, WiFi, GPS, etc., followed by testing their contacts, and repeating this procedure.
The problem is that such strategies are myopic and do not efficiently use the testing resources. This is
especially the case with COVID-19, where symptoms may show up several days after the infection (or
not at all, there is evidence to suggest that many COVID-19 carriers are asymptotic, but may spread the
virus). Such greedy strategies, miss out population areas where the virus may be dormant and flare up in
the future.

In this paper, we show that the testing problem can be cast as a sequential learning-based resource
allocation problem with constraints, where the input to the problem is provided by a time-varying social
contact graph obtained through various contact tracing tools. We then develop efficient learning
strategies that minimize the number of infected individuals. These strategies are based on policy iteration and look-ahead rules. We investigate fundamental performance
bounds, and ensure that our solution is robust to errors in the input graph as well as in the tests
themselves.

\vspace{.5cm}
\noindent \emph{Keywords:} Partially Observable Markov Decision Process, Approximation Algorithms, Sequential
Learning.
\end{abstract}

\section{Introduction}
The outbreak of the novel coronavirus (COVID-19) is unfolding as a major international crisis whose influence extends to every aspect of our daily lives. Testing is critical in identifying patients and carriers. Effective testing allows infected individuals to be quarantined, thus reducing the spread of COVID-19, \emph{saving countless lives, and helping to restart the economy safely and securely \cite{smart-testing,workshop,cdc-guidelines,testing-tracing}}. Testing capacity will remain a constraint for the foreseeable future. This means that we need to develop highly efficient testing strategies that make optimal use of our testing resources in order to minimize the number of infected individuals. 

These testing strategies can be greatly aided by contact tracing that provides health care providers information about the whereabouts of infected patients in order to determine whom to test. There have been significant efforts to improve contact tracing by developing apps that leverage the ubiquity of smart phones to automatically detect contacts between individuals within a pre-determined distance from each other (e.g., within 6 feet), the time duration of the contact, etc. \cite{gurman2020apple, wikiapps}. Countries that have been more successful in corralling the virus typically use a \emph{test, treat, trace, test} strategy that begins with testing individuals with symptoms, traces contacts of positively tested individuals via a combinations of patient memory, apps, WiFi, GPS, etc., followed by testing their contacts, and repeating this procedure. The problem is that such strategies are \emph{myopic and greedy} and do not efficiently use the testing resources. This is especially the case with COVID-19, where symptoms may show up several days after the infection (or not at all, there is evidence to suggest that many COVID-19 carriers are asymptotic, but may spread the virus) \cite{heneghan2020covid}. Such greedy strategies, often referred to as ``exploitation'' rules in the learning theory, miss out population areas where the virus may be dormant and flare up in the future. 

In this paper, we show that the testing problem can be formally cast as a \emph{sequential learning-based resource allocation problem with constraints}, where the input to the problem is provided by a time-varying social contact graph obtained through various contact tracing tools. Our goal is to develop efficient learning strategies that appropriately balance exploitation (testing high confidence individuals) as well as exploration (testing lower confidence individuals to identify potential unexplored areas, e.g., using group testing) to minimize the number of infected individuals. We will investigate fundamental performance bounds, and ensure that our solution is robust to errors in the input graph as well as in the tests themselves. 

\section{A Partially Observable Markov Decision Process Model with Contact Graph}\label{sec:pomdp}
We formulate the problem of sequential testing for COVID-19 as a Partially Observable Markov Decision Process (POMDP). The system of interest consists of $N$ individuals and evolves in discrete time $t\in [1,T]$. Let $X_i(t)\in\{0,1\}$ denote the ``hidden'' state of individual $i$ at $t$, where $X_i(t)=0$ means that $i$ is free of disease at $t$ and $X_i(t)=1$ indicates that $i$ is infected. We use the vector $X(t):=\left(X_1(t),X_2(t),\ldots,X_N(t)\right)\in \{0,1\}^{N}$ to represent the state of the entire system. Let $\cX:= \{0,1\}^{N}$ denote the state-space of the network. Note that the state vector $X(t)$ is never fully revealed to the learner\footnote{So the setup involves a partially observable MDP (POMDP), which is non-trivial to solve in general case.}.

\vspace{.5cm}

\noindent{\bf Test and Quarantine:} At each time $t\in[1,T]$, the learner has a unit budget to choose an individual $i\in [1,N]$ in order to ``sample'' (test for infection). Sampling an individual $i$ at $t$ reveals the state $X_i(t)$. We let $U(t)\in[0,N]$ denote the sampling decision at time $t$. In case no one is sampled at $t$, we let $U(t)=0$. The observation at $t$ is denoted $Y(t)$, and is given by $Y(t):=X_{U(t)}(t), t\in [1,T]. $
Note that if $U(t)$ is $0$, we assume $Y(t)$ to be deterministic, and hence reveals no information.
If sampled individuals are found to be infected, then they are ``quarantined,'' hence cannot spread the disease to their neighbors. We let $Q(t)$ denote the set of quarantined individuals until $t$. 

\vspace{.5cm}

\noindent{\bf Contact Graph:} The COVID-19 is spread by social contacts. We model the social contacts as a time-varying, weighted and undirected graph $G_t$ over a fixed node set $\cV=\{1,\ldots,N\}$, which denotes the $N$ individuals, i.e., $G_t=(\cV, \cE_t, w_t)$. Edges in graph correspond to social contacts and weights measure the extent of social contacts (e.g., contact duration, contact distance, number of times of contact, etc.). The social contact graph $G_t$ could be obtained from a combination of mobile apps, GPS/WiFi data, patient memory, etc. Note that the length of each time slot we considered in this testing system could be as small as seconds/minutes. Hence the graph $G_t$ could be highly dynamic or piece-wisely static, depending on the data updating frequency of the mobile app. 

\vspace{.5cm}

\noindent{\bf Active Edge:} In order to provide a unified framework for different sources of contact graph, and simplify exposition, we assume that only one single edge in the graph is \emph{active} at any time $t$, denoted as $\ell(t)$. 
Let $\cV'_t = \cV-Q(t)$ be those individuals that are not quarantined at $t$ (and hence ``free''), and $G'_t=(\cV'_t,\cE'_t,w'_t)$ be the vertex-induced subgraph of $G_t$. At each time $t$, the active edge $\ell(t)$ is sampled from the social contact subgraph $G'_t$ according to the edge weights. This means that the number of active social contacts are reduced as more and more confirmed cases are quarantined. We assume that $\ell(t)$ is revealed to the learner. Given $\ell(t) = (i,j)$, individuals $i$ and $j$ ``share" the disease with probability $p$, i.e., both of them become infected at time $t+1$ with a probability $p$ if either one of them was infected at time $t$. We will assume that the infection transmission probability $p$ is known. The case when $p$ is unknown, and needs to be learnt is considered separately.  

\vspace{.5cm}

\noindent{\bf State Transition:} Let us now look at the \emph{controlled} transition probabilities of the controlled Markov process $X(t)$. We first introduce some notations. For $x,y\in\cX$, define
\begin{align}
\Delta_1(x,y) = \mathbbm{1}\left\{\sum_{i=1}^{N} |x_i - y_i| = 1\right\}
\end{align}
and 
\begin{align}
\Delta_2(x,y) = 
\begin{cases}
i \mbox{ if } x_i\neq y_i \mbox{ and } \Delta_1(x,y)=1,\\
\emptyset \mbox{ otherwise }.
\end{cases}
\end{align}

Clearly, $\Delta_1(y,x)$ assumes value $1$ only if $x$ and $y$ differ in a single position. Since in our model we explicitly assume that the disease can spread to only one more person during two consecutive times, this function is $0$ if $x$ cannot evolve to $y$ in one single time-step. $\Delta_2$ provides us the node that ``transitioned'' to the diseased state when the system evolved in a unit step from $x$ to $y$.
Thus, the single-step controlled transition probability associated with the process $X(t)$ can be written as follows,
\begin{align}\label{eq:ptm}
P_t(x,y) = \Delta_1(x,y)p\sum_{i\in\cV'_t}w'_t(i,\Delta_2(y,x)).
\end{align}

\noindent{\bf Objective:} Let $\cF_{t}:=\cup_{s=1}^{t} \left(U(s),Y(s),\ell(s)\right)$ be the observation history of the learner. Then, the policy $\pi$ is a sampling decision at $t$ on the basis of $\cF_{t-1}$, i.e., $\pi:\cF_{t-1}\mapsto U(t),~t\in[1,T]$. Our goal is to find a policy that solves the following problem,
\begin{align}
&\min_{\pi}~\bE_{\pi}\left(\sum_{t=1}^{T} \|X(t)\|_{1}    \right) \label{def:covid_obj} \\
&~\mbox{ s.t. } \bE_{\pi}\left(\sum_{t=1}^{T} \mathbbm{1}\left( U(t)\neq 0\right) \right)\le C,\label{def:covid_cap}
\end{align} 
where $\|\cdot\|_{1}$ denotes the $L_1$ norm and $C$ is the total testing-capacity. The instantaneous cost $\|X(t)\|_{1}$ encourages the policy to keep the total number of infected individuals as low as possible, in an as early as possible manner. The capacity constraint~\eqref{def:covid_cap} is crucial because not many testing-kits are available during epidemics. An alternative, somewhat equivalent and simpler objective is to remove the capacity constraints altogther and include a cost for using testing-kits, 
\begin{align}\label{def:covid_modif}
\min_{\pi}\bE_{\pi}\left(\sum_{t=1}^{T} \|X(t)\|_{1} +\lambda \mathbbm{1}\left( U(t) \neq 0 \right)    \right),
\end{align}
where $\lambda>0$. In the remaining discussion, we restrict ourselves to~\eqref{def:covid_modif}. 
\begin{remark}
A natural but incorrect objective is to find a $\pi$ that maximizes the number of infections detected, i.e.,
\begin{align}\label{def:covid_problem}
\min_{\pi}\bE_{\pi}\left(-\sum_{t=1}^{T} Y(t)    \right)~~~ \mbox{ s.t. } \bE_{\pi}\left(\sum_{t=1}^{T} \mathbbm{1}\left( U(t)\neq 0\right) \right)\le C.
\end{align} 
However, we highlight the following issue with the formulation~\eqref{def:covid_problem}: the policy\slash algorithm is rewarded for catching as many infections as possible. We also note that the policy also does affect the evolution of the \emph{global state} $X(t)$. This is done by controlling the links $\ell(t)$ indirectly by quarantining those individuals whose tests turn out to be positive (recall that an infected person is quarantined, and is then not allowed to form links with any other person in the network). Hence, the objective~\eqref{def:covid_problem} encourages the development of a policy to infect as many people as possible (so that it can, at later stages, catch these cases). This dual affect of control~\cite{feldbaum1960dual} is clearly not desirable.
\end{remark}

\vspace{.5cm}

\noindent{\bf Belief State MDP Formulation}: 
We now introduce a belief state, which is the posterior distribution of $X(t)$ over the state space $\cX$. This allows us to transform the POMDP to a continuous-state MDP that involves evolution of the belief state. We denote the belief state by $\cI(t)=\{\cI(t,x)\}_{x\in\cX}$, where 
\begin{align*}
\cI(t,x):= \bP\left(X(t)=x| \cF_t \right),
\end{align*}
denotes the conditional probability associated with the system state equal to $x$. $\cI_{t}(x)$ can be computed recursively by utilizing the Bayes' Rule,
\begin{align}\label{eq:bayes}
\cI_{t+1}(x) &= \sum_{y\in\cX} \cI_{t}(y)\bP\left( Y_{U(t)}  |X(t)=y \right)P_t(y,x),
\end{align}
where the state transition probabilities $P_t(y,x)$ are as discussed in~\eqref{eq:ptm}.

\vspace{.5cm}

\noindent{\bf Optimal Policy:} The sampling policy that is optimal for the problem~\eqref{def:covid_obj}-\eqref{def:covid_cap} can be obtained by solving the following set of non-linear Dynamic Programming equations~\cite{krishnamurthy2016partially},
\begin{align}
V_{t}(\cI_{t}) &= \sum_{x\in \cX}\|x\|_1\cI_t(x) + \min_{u \in [0,N]}\left( \bE~V_{t+1}(\cI_{t+1}) +\lambda\mathbbm{1}\{u\neq 0\}  \right),\label{eq:bellman_1}\\
V_{T}(\cI) &= \sum_{x\in \cX}\|x\|_1\cI(x),~\forall \cI\in \Delta(\cX),\label{eq:bellman_2}
\end{align}
where $\Delta(\cX)$ denotes simplex on $\cX$, $\cI_{t}$ denotes representative belief state at time $t$, and the function $V_t(\cdot)$ denotes the value function at time $t$. Optimal sampling action at time $t$ in state $\cI_t$ corresponds to minimizer of r.h.s. in the above equation.~Equations~\eqref{eq:bellman_1},~\eqref{eq:bellman_2} are computationally intractable as $\Omega(2^N)$. Thus, we propose tractable provably approximate solutions next. 

\subsection{Provably Sub-optimal Value Iteration Approximation}\label{sec:prove_subopt}
We describe an approximation method with low computational complexity for the POMDP~\eqref{def:covid_modif}. Despite~\eqref{def:covid_modif} being a continuous-state MDP, it has a finite dimensional characterization~\cite{smallwood1973optimal,sondik1978optimal}. This characterization is exploited in~\cite{monahan1982state,cassandra1994acting,lovejoy1991computationally} in order to develop approximate solutions that are computationally tractable. Among these approaches, \cite{lovejoy1991computationally}  provides upper and lower bounds to the proposed approximation scheme, and hence also has theoretical guarantees. The following result is taken from~\cite{lovejoy1991computationally}.
\begin{theorem}\label{th:sondik_lovejoy}
Consider the Bellman equations~\eqref{eq:bellman_1},~\eqref{eq:bellman_2}, the associated value functions $V_{t}(\cdot),t\in[1,T]$ and the optimal policy $\pi\ust=\left\{\pi_t\ust(\cI_t)\right\}_{t=1}^{T}$. They have the following finite-dimensional characterization.
\begin{enumerate}
\item $V_t(\cI)$ is piecewise-linear and concave with respect to $\cI$. Thus, $V_t(\cI) = \min_{\gamma\in\Gamma_k} \gamma^{T} \cI$, for any $t\in[1,T]$, where $\Gamma_t$ is a finite set of $\cX$ dimensional vectors.
\item $\pi\ust_t(\cI_t)$ has the following finite dimensional characterization: The belief space $\Delta(\cX)$ can be partitioned into at most $|\Gamma_t|$ convex polytopes. In each
such polytope, the optimal policy $\pi\ust_t(\cI_t)$ is a constant corresponding to a single action. 
\end{enumerate}
\end{theorem}

\vspace{.5cm}

Since the sets $|\Gamma_t|$ can be quite large, we can reduce the computational cost from $\Omega(2^N)$ to $O(\text{poly}(N))$ by cleverly choosing ``approximation sets" $\hat{\Gamma}_t$ having small cardinalities. The resulting ``approximate value function" $\bar{V}_t(\cdot)$ would then yield an approximately optimal policy. This is the basis of~\cite{lovejoy1991computationally}'s approximation scheme that is stated below, which gives an upper-bound to the true value functions $V_t(\cdot)$.

\begin{itemize}
\item \textbf{Initialize}: $\hat{\Gamma}_T = \Gamma_T= \{c_T\}$, where $c_T$ is the terminal cost vector.

\vspace{.5cm}

\item \textbf{Step 1.} Given a set of vectors $\Gamma_t$, construct the set $\bar{\Gamma}_t$  by pruning $\Gamma_t$ as follows: Pick any $R$ belief states $i_1,i_2,\ldots,i_R$ in the belief simplex $\Delta(\cX)$. \footnote{Any homotopy algorithm for solving equations without special structure, and which uses Freudenthal Triangulation  can be used for this step. Interested readers see~\cite{munkres2016elementary} for technical terms.} Then perform the following operations,

\begin{align*}
\bar{\Gamma}_t = \left\{\arg\min_{\gamma\in\Gamma_t}  \gamma^{T} i_r, r=1,2,\ldots,R \right\}.
\end{align*}

\vspace{.5cm}

\item \textbf{Step 2.} With $\bar{\Gamma}_t$, obtain $\Gamma_{t-1}$ by using any standard POMDP algorithm.

\vspace{.5cm}

\item \textbf{Step 3.} $t\rightarrow t-1$ and goto \textbf{Step 1}.
\end{itemize}

\begin{figure}
\centering
  \includegraphics[width=0.7\linewidth]{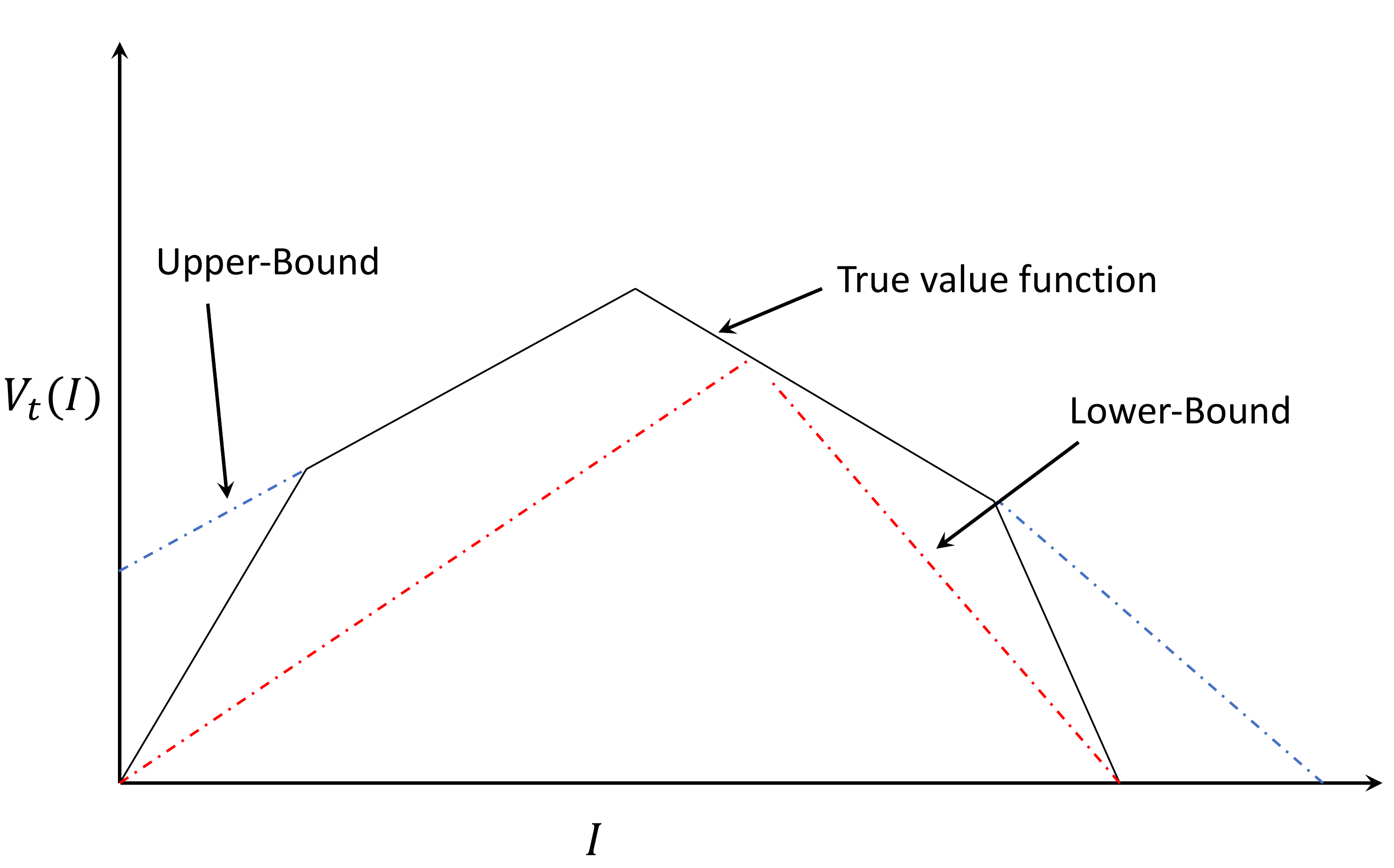}
  \caption{The true value function $V_t(\cdot)$ (black) is ``sandwiched" between the upper (blue) and lower (red) bound curves. Note that upper-bound and $V_t(\cdot)$ are equal within an interval.}
  \label{fig:lovejoy}
  \end{figure}

To get a lower-bound, choose any $R$ belief states $\{i_j\}_{j=1}^{R}$ and construct a linear interpolation between the points $\left(i_j,V_t(i_j)\right)$. It then follows from the concavity of $V_t(\cdot)$, that the resulting curve lies below $V_t(\cdot)$. Hence, the true value function $V_t(\cdot)$ is ``sandwiched" between the upper and lower bound, as depicted in Figure.~\ref{fig:lovejoy}.

\section{Provably Sub-Optimal Low Complexity Algorithms}
We now introduce two broad class of algorithms that are easy to implement, and we provide guarantees on their performance.
\subsection{Policy Iteration Approximation}
The idea is to begin with a naive sampling policy $\pi_0$ for which the value function is easily computable, and then employ one step of policy iteration in order to obtain a policy $\pi_{im}$ that is better than $\pi_0$. Since the policy iteration operator corresponds to Newton's method applied on the policy-space~\cite{whittle1996optimal}, a single application of policy iteration is supposed to yield vast improvements. More details regarding the ``convergence rates'' of such procedures can be found in~\cite{puterman2014markov}. We now give an example of an easily-computable $\pi_0$, and also describe the policy iteration technique.

\emph{Open-Loop Policy $\pi_0$}: At time $t=0$ the user picks $T$ nodes out of $N$ nodes, arranges them in some order and decides to sample them according to this order. These $T$ nodes, say $i_1,i_2,\ldots,i_T$, are then sampled during the next $T$ time-slots. This policy is clearly an \emph{open-loop} policy since it makes decisions in a non-adaptive manner, i.e., it does not change its decision regarding which node to sample despite gaining more information during the experiment. The value function corresponding to $\pi_0$ can be obtained by solving the following set of equations,
\begin{align}
V_{t,\pi_0}(\cI_{t}) &= \sum_{x\in \cX}\|x\|_1\cI_t(x) + \left( \bE~V_{t+1}(\cI_{t+1}) +\lambda\mathbbm{1}\{i_t\neq 0\}  \right),\\
V_{T,\pi_0}(\cI) &= \sum_{x\in \cX}\|x\|_1\cI(x),~\forall \cI\in \Delta(\cX),
\end{align}
where $\cI_{t+1}$ is calculated from~\eqref{eq:bayes} with $U_t = i_t$, and the sub-script $\pi_0$ denotes that the value function is associated with the policy $\pi_0$. 

\emph{Policy Iteration}: The sampling decision $U(t)$ at time $t$ is obtained by solving the following equation
\begin{align}\label{eq:pol_iter}
 \min_{u \in [0,N]} \bE\left(~V_{t+1,\pi_0}(\cI_{t+1}) +\lambda\mathbbm{1}\{u\neq 0\} \right),~t\in[0,T-1].
\end{align}

We summarize the discussion of this section as the following result.

\vspace{.5cm}

\begin{theorem}
Consider the sampling policy $\pi_{im}$ which makes decisions as in~\eqref{eq:pol_iter}, and is obtained by utilizing a single step of the policy improvement operator upon the policy $\pi_0$. We then have that $\pi_{im}$ yields a better performance than $\pi_0$, i.e., their value functions satisfy $V_{t,im}(\cdot) \le V_{t,\pi_0}(\cdot)$.  
\end{theorem}

\subsection{Cost-to-go-Approximations via Look-Ahead Rules}\label{sec:lar}
The idea behind this approach is that instead of solving the Dynamic Programming equations~\eqref{eq:bellman_1},~\eqref{eq:bellman_2} exactly, we derive only an approximation $\tilde{V}_t$ of the true value functions $V_t$. Such approximations yield more computationally tractable approaches, but yield only a suboptimal policy. There are many approaches to derive such approximations, however we will restrict ourselves to \emph{look-ahead rules}~\cite{bertsekas2005dynamic,bertsekas1995dynamic}. Another approach yields an \emph{index rule} that attaches an index to each ``arm'' (an individual), and then samples the individual with the largest value of index. Some examples of such index rules, and more details on how to derive these policies can be found in~\cite{Whittle2011Book,kadota2016minimizing,kadota2018minimizing,guo2016index,guo2015optimal,singh2015index}. We next discuss the look-ahead approach.

Let $\tilde{V}_{t+1}$ be an approximation of the value function at time $t+1$. If $\cI_{t+1}$ denotes the belief state at time $t+1$, then the decision at time $t$ is obtained by solving the following optimization problem,
\begin{align}\label{label:lar}
u_t \in \arg\min_{u \in [0,N]}\left( \bE~\tilde{V}_{t+1}(\cI_{t+1}) +\lambda\mathbbm{1}\{u\neq 0\}  \right).
\end{align}
We will make the following assumption in order to analyze the performance of look-ahead rules.
\begin{assumption}\label{assum:lar}
For all $x$ and times $t\in [1,T]$, we have that
\begin{align*}
\tilde{V}_t(x) \ge \sum_{x\in \cX}\|x\|_1\cI_t(x) + \min_{u \in [0,N]}\left( \bE~\tilde{V}_{t+1}(\cI_{t+1}) +\lambda\mathbbm{1}\{u\neq 0\}  \right),
\end{align*}
where in the above $x$ denotes a representative state ($N$ dimensional vector comprising of $0$s and $1$ s). 
\end{assumption}
Under the above assumption, we can prove the following appealing property of the look-ahead policy. 
\begin{theorem}
Consider the problem of designing an efficient sampling procedure for testing individuals for disease. Let $V_{t,la}(\cdot)$ denote the cost-to-go function of the look-ahead policy that makes decisions according to~\eqref{label:lar}. Also, let Assumption~\ref{assum:lar} be satisfied. We then have that 
\begin{align*}
V_{t,la}(\cI_{t}) \le \sum_{x\in \cX}\|x\|_1\cI_t(x) + \min_{u \in [0,N]}\left( \bE~\tilde{V}_{t+1}(\cI_{t+1}) +\lambda\mathbbm{1}\{u\neq 0\}  \right). 
\end{align*}
\end{theorem}
We now provide some examples of such look-ahead rules.

We begin with a simpler problem in which we only have to make sampling decision for only a single time-step\slash resource. In this case, a greedy policy makes a sampling decision that minimizes the instantaneous cost, as follows 
\begin{align}
\pi^{greedy}(\cI_t)\in\arg\min_{u\in[0,N]} \left(\sum_{x\in A(t)}\cI_t(x)\right) \left[p\sum_{i\in\cV}w'_t(u,i)\right]+\lambda\mathbbm{1}\{u\neq 0\},\label{def:greedy_1}
\end{align}
where $A(t,u) := \left\{x\in \cX: x_u = 1, u\notin Q(t)  \right\}$. The set $A(t,u)$ represents those possibilities in which user $u$ is infected and not quarantined. Let $V^{greedy}(\cdot)$ denote the value function for this greedy rule, which is 
\begin{align*}
V^{greedy}(\cI_t) = \sum_{x\in \cX}\|x\|_1\cI_t(x) + \bE\left[\sum_{x\in \cX}\|x\|_1\cI_{t+1}(x) + \lambda\mathbbm{1}\{U\neq 0\}\Big| U=\pi^{greedy}(\cI_t) \right].
\end{align*}
Note that the greedy policy $\pi^{greedy}$ is an \emph{exploitation-only policy} as it will only test individuals with high confidence to be infected. This is not a good policy as it \emph{does not explore} controlling the virus as early as possible. To introduce a certain level of exploration, we now apply one step of policy improvement, i.e., the Bellman operator~\eqref{eq:bellman_1},~\eqref{eq:bellman_2} to the greedy policy. The resulting policy generates the sampling decisions $U(t)$ as follows,
\begin{align*}
U(t) \in \arg\min_{u \in [0,N]} \bE~V^{greedy}(\cI_{t+1}).
\end{align*}
This resulting policy is one-step look ahead policy, and denote it $\pi^{os}$. Note that the computational complexity of $\pi^{os}$ is $O(\text{poly}(N))$.

\vspace{.5cm}

\section{More Complex Environments}\label{sec:dini}
The models considered in the previous sections are too simplistic, and may not be adequate to capture many real-world scenarios. In this section, we briefly discuss how to enhance these models and algorithms in order to provide solutions for more complex scenarios. 
\subsection{Group Testing}
One way to improve the efficiency of resource allocation in~\eqref{def:covid_modif} is to employ group testing~\cite{du2000combinatorial}. In this procedure, samples of multiple individuals are combined into a single ``mixture" sample and tests performed on the mixture. In case the result is negative,  all the component individuals are declared negative, which is especially useful when testing lower confidence population areas (e.g., during exploration). However, if the test is positive,  a subset of these individuals  are carefully selected for conducting further tests, allowing identification of all positive individuals. Such a procedure generally saves the number of tests required, and is immensely useful during testing-kit shortages. Group testing can be incorporated into the model of Section~\ref{sec:pomdp} as follows. The cost incurred by the system remains the same as in~\eqref{def:covid_modif}. However, the action-space, i.e., the choice of controls is now all possible subsets of $\{1,2,\ldots,N\}$, i.e., $U(t)\in 2^{\cV}$ and denotes the set of individuals are to chosen for collective testing at $t$. We will seek to develop adaptive algorithms that perform group testing in an efficient manner and quantify the additional gains (in terms of the additional number of people that were prevented from getting infected) from group testing.

\subsection{Inaccurate Testing}
The model considered in Section~\ref{sec:pomdp} assumes that the testing result will reveal the current state of the tested individual. However, in practice, the tests are not 100\% accurate (e.g., PCR tests for COVID-19 have a high false negative). Our model can be readily extended to accommodate this noisy testing as one can apply the Bayes' rule on the observation to infer the current state of the tested individual. Yet, this introduces one interesting question when we deploy group testing. 
Here, note that although group testing itself introduces testing errors in false negatives if the samples get sufficiently diluted, it is also an efficient way to deal with testing error as one individual could be tested multiple times keeping the overall average sample much less than 1. We will study how to efficiently allocate group testing in order to reduce the overall testing error.
\subsection{Noisy Contact Graph}
The model considered in Section~\ref{sec:pomdp} assumes that the social contact graph are known and the COVID-19 is spread through these contacts. However, in practice, this is rarely the case because contact tracing only provides approximate coverage and noisy linkages. Hence, we need to extend the state transition kernel to allow some unknown source of infection $p_{u}$. Besides, it is hard to know the transmission probability $p$ given a contact a priori as well. We will extend the problem~\eqref{def:covid_modif} to the case without knowledge of $p$ and $p_{u}$. Such an optimization necessarily entails learning these unknown parameters. The algorithm thus has to perform a trade-off in which it makes sub-optimal choices for sampling people, which enables it to learn these parameters. We plan to develop learning algorithms that perform this trade-off in an optimal manner.
\subsection{Information Directed Sampling Approach for Learning}
The approximation methods we proposed in the previous Sections are computationally tractable compared to the optimal policy~\eqref{eq:bellman_1} and~\eqref{eq:bellman_2}. These methods are practical for moderate population $N$, e.g., a city. However, its $O(\text{poly}(N))$ computational complexity does not allow it to scale to large population $N$, e.g., nationwide. One promising approach to further reduce the computational complexity is to consider a compressed/kernelized policy space. This is reasonable for practical situations where tradeoffs are often made between optimality and feasibility. For example, in practice, we may only be able to play with the portion of the total testing budget that could be used to explore asymptomatic individuals. This motivates us to consider a class $\Pi$ of policies parametrized by parameter $\theta\in\Theta\subset R^{d}$. Our goal is to ``learn'' the best policy from amongst the class $\left\{\pi_{\theta}: \theta\in \Theta  \right\}$. One possibility is to employ Thompson sampling, or efficient Bayesian information collection type of learning rules, e.g.~\cite{srinivas2009gaussian,russo2014learning}. We briefly describe the approach below. Let $T_s$ denote a ``sufficiently'' large time-period. Total time horizon of $T$ steps is divided into ``episodes'' of $T_s$ slots each. We employ a fixed policy $\pi_{\theta(k)}$ in the $k$-th episode, that begins at time $\tau_k:= k~T_s$. The following optimization problem is solved at time $\tau_k$ in order to derive $\pi_{\theta(k)}$: 
$\max_{\theta\in\Theta}  \bar{V}(\pi_{\theta}) + \beta_{k} var(V(\pi_{\theta} )  )$,
where $V(\pi)$ is the performance of policy $\pi$ during $T_s$ time-steps. $V(\pi)$ is a random variable, because it depends upon unknown parameters, and $\bar{V}(\cdot), var(V(\cdot))$ denote its mean value and variance respectively. $\beta_k$ is a suitably chosen step-size, that converges to $0$ as $k\to\infty$. It will be interesting to characterize the performance of this learning rule, more specifically how it scales with $T$.    


\section{Future Works}
The approximation procedure of Section~\ref{sec:prove_subopt} lacks a characterization of the gap between the upper and lower bounds. This gap depends upon the user's choice of $R$, and the sample belief states $i_1,i_2,\ldots,i_R$. An important problem would be to characterize this dependence, thereby allowing us to make ``optimal'' choices for these hyperparameters. This would also provide us with a ``convergence'' rate for the approximation algorithm of~\cite{lovejoy1991computationally}, i.e., how fast the approximation error goes to $0$ as the granularity controlled by $R$, is increased.

The look-ahead policies have performance guarantees under certain conditions on the system parameters, which in our case translate to conditions on the social contact graph. It is often the case that $\pi^{os}$ is near-optimal since it ``looks into the future'' while making decisions. 
It is of interest to investigate the performance of $\pi^{os}$; more specifically to seek a characterization of its sub-optimality gap. 


\bibliographystyle{IEEEtran}
\bibliography{pomdp.bib}

\end{document}